\documentclass[journal]{IEEEtran}
\usepackage{graphicx}
\usepackage{subfigure}
\usepackage{url}
\usepackage{times}
\usepackage{amssymb}
\usepackage{array}
\usepackage[ruled,lined]{algorithm2e}
\usepackage{amsmath}
\usepackage{booktabs}
\usepackage{multirow}

\begin{document}

%%%%%%%%% TITLE
\title{Enhancing the Discriminative Feature Learning \\ for Visible-Thermal Cross-Modality Person Re-Identification}

\author{Haijun Liu and Jian Cheng

\thanks{H. Liu and J. Cheng are with the School of Information and Communication Engineering, University of Electronic Science and Technology of China, Chengdu, Sichuan, China, 611731.(haijun\_liu@126.com)}}

\maketitle

%%%%%%%%% ABSTRACT
\begin{abstract}
  Existing person re-identification has achieved great progress in the visible domain, capturing all the person images with visible cameras. However, in a 24-hour intelligent surveillance system, the visible cameras may be noneffective at night. In this situation, thermal cameras are the best supplemental components, which capture images without depending on visible light. Therefore, in this paper, we investigate the visible-thermal cross-modality person re-identification (VT Re-ID) problem. In VT Re-ID, there are two knotty problems should be well handled, cross-modality discrepancy and intra-modality variations. To address these two issues, we propose focusing on enhancing the discriminative feature learning (EDFL)  with two extreme simple means from two core aspects, (1)  skip-connection for mid-level features incorporation to improve the person features with more discriminability and robustness, and (2) dual-modality triplet loss to guide the training procedures by simultaneously considering the cross-modality discrepancy and intra-modality variations. Additionally, the two-stream CNN structure is adopted to learn the multi-modality sharable person features. The experimental results on two datasets show that our proposed EDFL approach distinctly outperforms state-of-the-art methods by large margins, demonstrating the effectiveness of our EDFL to enhance the discriminative feature learning for VT Re-ID.
\end{abstract}

%%%%%%%%% BODY TEXT
\section{Introduction}
\label{sec:intro}
Person re-identification (Re-ID), aiming at solving the problem of retrieving a person of interest across multi non-overlapping cameras deployed at different locations, has received increasing interests in computer vision community due to its importance in intelligent video surveillance \cite{zheng2016person, karanam2018systematic,chen2018person}.
Almost all of the current Re-ID models are focusing on the visible-visible person images matching, the most common single-modality Re-ID task, i.e., given a probe person image (or video) and match it against a set of gallery images (or videos) captured by other disjoint cameras, where all data are obtained from the visible cameras. Under the single visible modality, encouraging performances of visible-visible person Re-ID (VV Re-ID) have been achieved \cite{liu2017hydraplus,su2017pose,wang2018learning,li2018harmonious,chang2018multi,wang2018mancs,liu2019gallery}.

However, in a 24-hour intelligent surveillance system, only the visible cameras are not enough, especially when light is poor or unavailable (e.g. during the night). The visible cameras can not capture the appearance information of persons. In this case, image capturing devices without depending on visible light are necessary, such as thermal cameras or depth cameras.
Recently, the depth images are always captured by the RGB-D cameras (e.g. Kinect), nevertheless, which are rarely deployed in practical surveillance system  because they are expensive, always used in door and imaging with distance limitations.
Comparatively, the thermal cameras, using the infrared light to capture the persons, are commonly used in practical video surveillance systems. Moreover, most surveillance cameras can automatically switch between visible and thermal modes according to the light conditions.

\begin{figure}
\centering
\includegraphics[width=85mm]{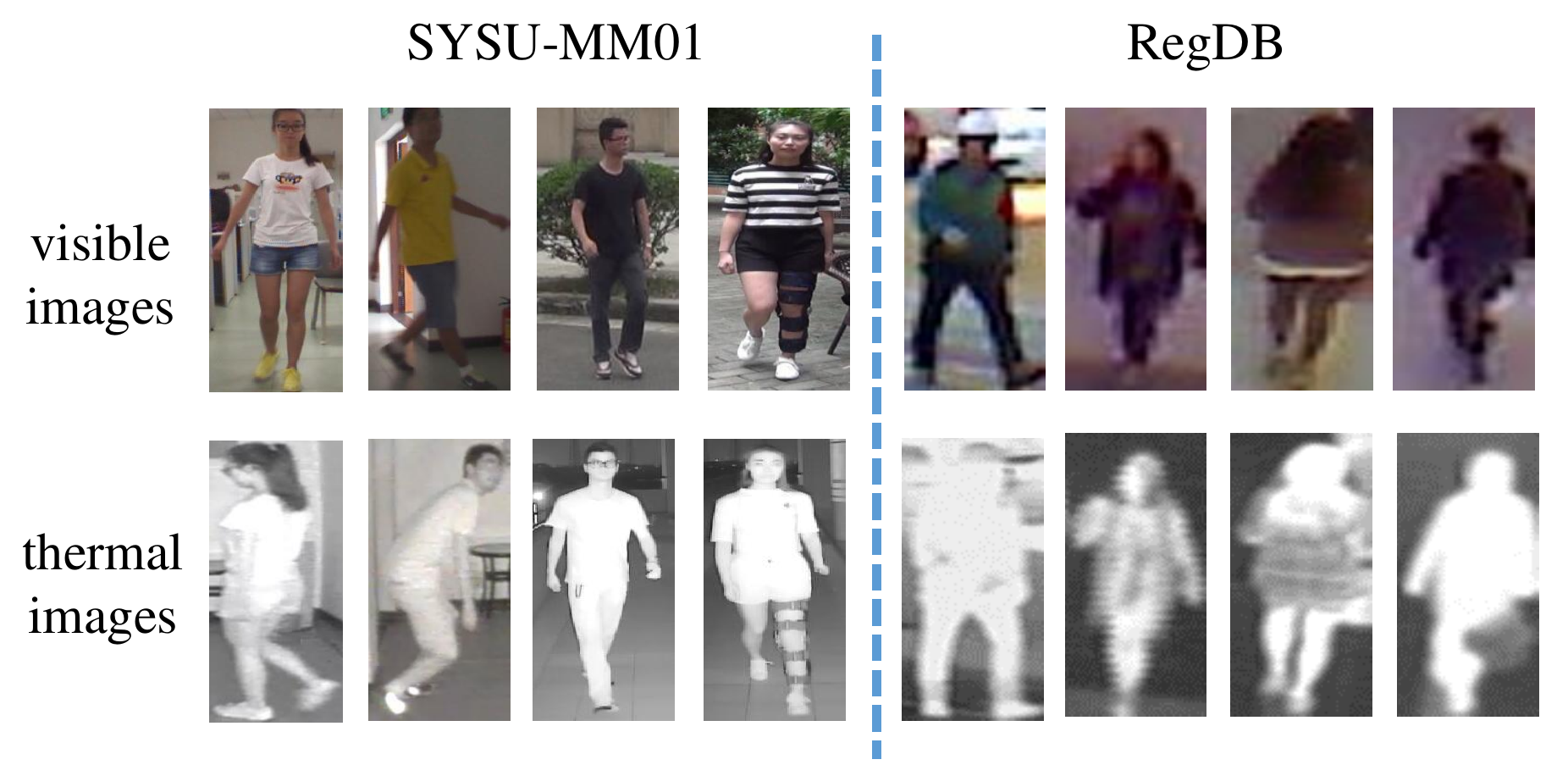}
\caption{Illustration of the visible-thermal images, from two datasets SYSU-MM01 \cite{wu2017rgb} and RegDB \cite{nguyen2017person}, for cross-modality person re-identification. The first row is the visible images, while the second is the thermal images. Each column contains the images from the same person. }
\label{fig:image_illus}
\end{figure}

In practical scenarios, a 24-hour intelligent surveillance system, the probe image may be obtained from the visible cameras during the daytime, while the gallery images may be captured by the thermal cameras during the nighttime, where images are from different modalities.
Therefore, it is necessary to study the cross-modality person Re-ID problem. In this paper, we address the visible-thermal person Re-Identification (VT Re-ID) in a 24-hour intelligent surveillance system, which is also termed as RGB-infrared person Re-ID \cite{wu2017rgb}.

For the VT Re-ID task, there are two big problems should be paid much attention. On the one hand, similar to other cross-modality recognition tasks (heterogenous face recognition \cite{reale2017deep,he2017learning,wu2018coupled,deng2019mutual} and text-to-image retrieval \cite{cao2017collective,cornia2018towards,wang2019learning}), the large \textbf{cross-modality discrepancy} is the biggest issue for VT Re-ID, which is arisen from the different reflective visible spectrums and sensed emissivities of visible and thermal cameras. As shown in Figure \ref{fig:image_illus}, the visible images in the first row have three channels containing sufficient color information of person appearance, while the thermal images in the second row have one channel containing information of in-visible light. This leads the color information, which is the most important appearance cue for identifying person in the VV Re-ID, can hardly be used on the heterogeneous data for VT Re-ID.
On the other hand, similar to traditional VV Re-ID, the large \textbf{intra-modality variations} caused by viewpoint changing and different human poses, also brings difficulties to VT Re-ID, resulting to a much more challenging problem.

\begin{figure}
\centering
\includegraphics[width=85mm]{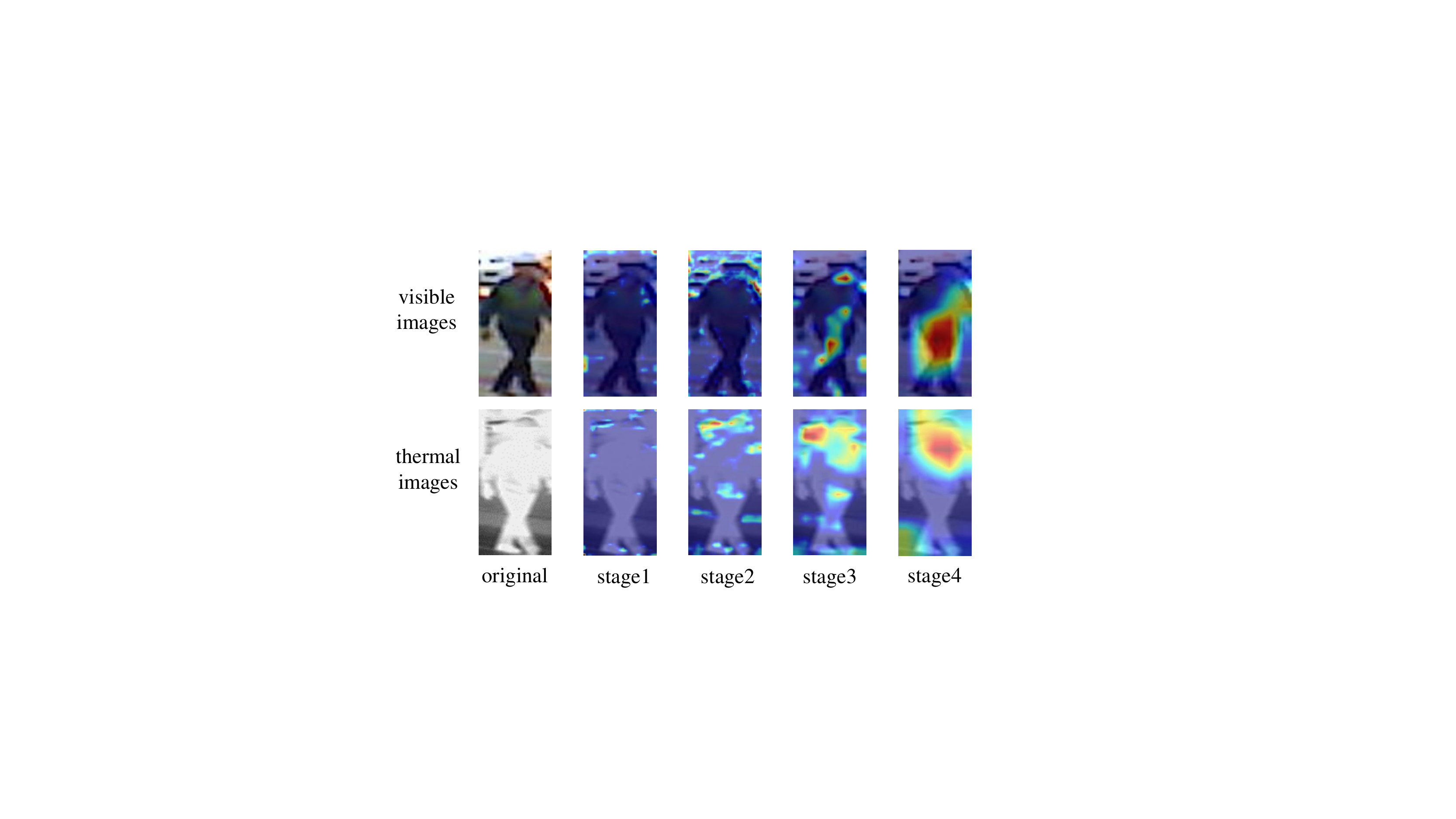}
\caption{The location visualization of different stages of ResNet50 \cite{he2016deep} model for visible-thermal person Re-ID. The Grad-CAM method \cite{selvaraju2017grad} is respectively performed on two ResNet50 models, the backbone networks in our baseline, for visible and thermal images.}
\label{fig:grad_cam}
\end{figure}

For the aforementioned two issues, cross-modality discrepancy and intra-modality variations, we try to address them by focusing on enhancing the discriminative feature learning from the following aspects.
(1) Two-stream convolutional neural network (CNN) structure. Two-stream CNN is adopted to extract the person features for VT Re-ID, which includes a visible stream and a thermal steam. Two independent CNNs are firstly utilized to learn the modality-specific information, addressing the cross-modality discrepancy problem. Then some shared layers are further utilized to embed these modality-specific information into a common space. The two-stream CNN structure can generate the multi-modality sharable features by simultaneously considering the modality commonality and discrepancy.
(2) Mid-level features incorporation. It is well known that, different CNN feature maps correspond to different semantic levels, from bottom to top layers, tending to visual concepts that are of higher semantic level and more abstract \cite{zeiler2014visualizing}. Moreover, as shown in Figure \ref{fig:grad_cam}, through the Grad-CAM method \cite{selvaraju2017grad}, which is a technique for visual explanations of deep networks via gradient-based localization, we can find that different layers of deep CNN truly focus on different location for predicting the concepts. Therefore, we try to incorporate the mid-level features from middle layers of the CNN model through skip-connection to enhance the person features with more discriminability and robustness.
(3) Dual-modality triplet loss. We design a novel bi-directional dual-modality triplet loss to guide the training procedures. Triplet loss \cite{hermans2017defense} is respectively adopted on both of the cross modalities and intra modality, to make those person features from the same identity close to each other, while those person features from different identities further away, no matter the person images are within which modality. The dual-modality triplet loss can guide the training procedures to enhance the discriminative person feature learning by simultaneously considering the cross-modality discrepancy and intra-modality variations.

In summary, our paper has the following contributions.
\begin{itemize}
\item Mid-level features incorporation is adopted through skip-connection for VT Re-ID to enhance the person features with more discriminability and robustness.
\item Dual-modality triplet loss is adopted to constrain on both of the cross-modality and intra-modality, to address the cross-modality discrepancy and intra-modality variations.
\item Our proposed enhancing the discriminative feature learning (EDFL) method, in an end-to-end manner with the two-stream CNN structure, outperforms the state-of-the-art approaches by large margins on two public VT Re-ID datasets.
\end{itemize}

\section{Related work}
\label{sec:relatedwork}
To our best knowledge, in the literature there are only five pioneer works about visible-thermal person Re-ID.

\noindent \textbf{RGB-Infrared Cross-Modality Person Re-Identification \cite{wu2017rgb}.} Wu et al. firstly proposed to address the the RGB-IR cross-modality Re-ID problem and contributed a new multiple modality person Re-ID dataset named SYSU-MM01. Moreover, they further proposed deep zero-padding method for training one-stream network towards automatically evolving domain-specific nodes in the network for cross-modality matching.

\begin{figure*}
\centering
\includegraphics[width=17cm]{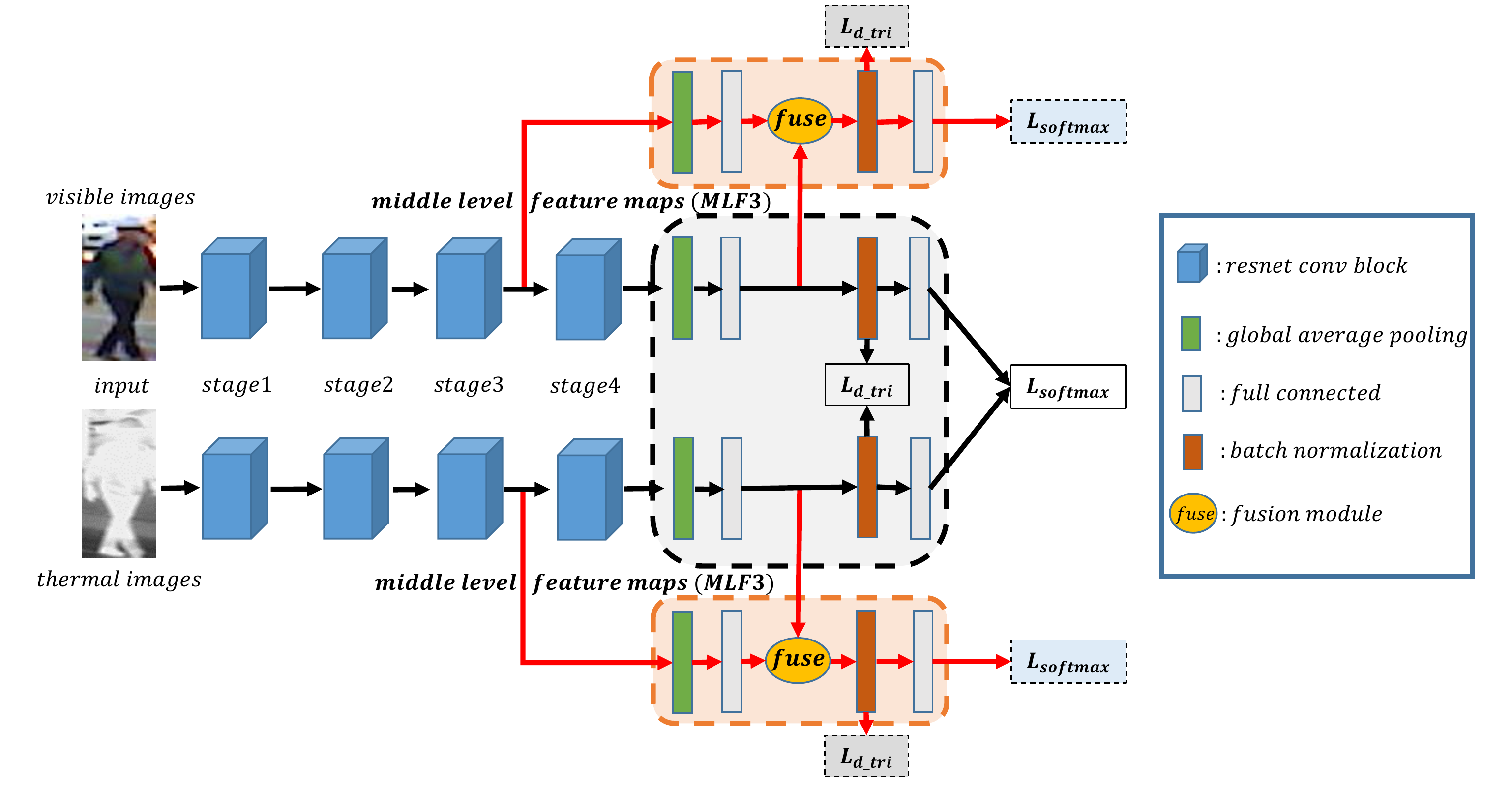}
\caption{The proposed EDFL framework for VT Re-ID. Two-stream CNN structure is adopted to extract person features, one stream for visible images and the other stream for thermal images. For example, we take two ResNet50 models as the backbone in each stream respectively, which are independent to each other to extract the modality-specific information. Then top full connected layers are followed in a parameters sharing manner for both two streams to embed multi-modality person features in a common feature space. Finally the proposed dual-modality triplet loss ($L_{d\_tri}$) and identity softmax loss ($L_{softmax}$) are adopted for network training. Moreover, the mid-level features from middle layers of CNN model are fused with the final feature of backbone to improve the discriminability of person features. The black lines without $L_{d\_tri}$ are the baseline procedures, while adding the red lines are procedures of skip connections for the mid-level features incorporation. Those layers in the dashed squares with the same color share the parameters.}
\label{fig:framework}
\end{figure*}

\noindent \textbf{Cross-Modality Person Re-Identification with Generative Adversarial Training \cite{dai2018cross}.} Dai et al. proposed a novel cross-modality generative adversarial network (termed cmGAN) to learn discriminative common representations. cmGAN consists of a deep convolutional neural network as generator for learning image representations and a modality classifier as discriminator which tries to discriminate between RGB and infrared image modalities.

\noindent \textbf{Hierarchical Discriminative Learning for Visible Thermal Person Re-Identification \cite{ye2018hierarchical}.} Ye et al. proposed a hierarchical cross-modality matching model for VT Re-ID by jointly optimizing the modality-specific and modality-shared metrics. The person features similarity metric are learnt in a hierarchical manner, firstly by modality-specific metric then by the modality-shared metric.
However, it performs the VT Re-ID task in two separate steps, two-stream CNN network for feature extracting and then hierarchical discriminative cross-modality metric learning (HCML) for similarity metric.

\noindent \textbf{Visible Thermal Person Re-Identification via Dual-Constrained Top-Ranking \cite{ye2018visible}.} Based on the aforementioned hierarchical discriminative learning method, Ye et al. then proposed a dual-path network with a novel bi-directional dual-constrained top-ranking (DCTR) loss to learn discriminative features. It mainly has two advantages: 1) end-to-end feature learning directly from the data without extra metric learning steps, 2) it simultaneously handles the cross-modality and intra-modality variations by the dual-constrained top-ranking loss to ensure the discriminability of the learnt person features.

\noindent \textbf{Learning to Reduce Dual-level Discrepancy for Infrared-Visible Person Re-identification \cite{wang2019learning1}.} Wang et al. proposed a novel dual-level discrepancy reduction learning (D$^2$RL) scheme to separately handle the two discrepancies: modality discrepancy and appearance discrepancy. For reducing the modality discrepancy, an image-level sub-network is trained to translate an infrared image into its visible counterpart and a visible image to its infrared version. Then with the image-level sub-network, we can unify the representations for images with different modalities. With the help of the unified multi-spectral images, a feature-level sub-network is trained to reduce the remaining appearance discrepancy through feature embedding. The two-level sub-networks can take their responsibilities cooperatively and attentively.

Our proposed approach is based on the framework of dual-constrained top-ranking (DCTR) \cite{ye2018visible}. However, our proposed approach is different from DCTR at least in the following two aspects. 1) We incorporate the mid-level features of CNN models through skip-connection to enhance the person features with more discriminability and robustness. 2) Our dual-modality triplet loss, modeling the cross-modality discrepancy and intra-modality variations in a hard triplets mining manner, is different from the dual-constrained top-ranking loss, modeling the cross-modality and intra-modality constraints in a contrastive top-ranking manner. By the two modifications, we can strongly improve the cross-modality visible-thermal person Re-ID performance.

\section{Our proposed method}
\label{sec:our_method}

In this section, we will introduce the framework of our proposed enhancing the discriminative feature learning (EDFL) model for VT Re-ID, as depicted in Figure \ref{fig:framework}. EDFL model mainly consists of three components: (1) the two-stream backbone architecture, (2) the skip-connection for mid-level features incorporation and (3) the loss, dual-modality triplet loss and identity softmax loss.

\subsection{Two-stream structure}
\label{ssec:backbone}
We adopt a two-stream structure network to extract the person features, including a visible stream and a thermal stream. In each stream, it can take any deep CNNs designed for image classification as the backbone, e.g., Google Inception \cite{szegedy2017inception} and ResNet \cite{he2016deep}. We will take the ResNet50 model as an example, with the consideration of its competitive performance in some Re-ID systems \cite{sun2018beyond,wang2018mancs} as well as its relatively concise architecture.
ResNet50 model mainly consists of four res-convolution blocks, $stage1$, $stage2$, $stage3$ and $stage4$, as illustrated in Figure \ref{fig:framework}. The res-convolution blocks are independent in the two streams, aiming to learn the modality-specific information, addressing the cross-modality discrepancy problem.

Then some shared layers are further utilized to embed these modality-specific information into a common space, learning a multi-modality sharable space to bridge the gap between two heterogenous modalities. As shown in Figure \ref{fig:framework}, those shared layers in the orange dashed square are as following.
(1) A full connected layer with output dimension $d$ is added after the global average pooling layer, as a bottleneck to reduce the dimensions if necessary.
(2) Then a batch normalization ($BN$) layer is added sequentially, whose output would be adopted to perform the metric learning with our proposed dual-modality triplet loss.
(3) Finally a full connected layer with desired dimensions (corresponding to the number of identities of person in our model) is adopted to perform the classification with softmax loss.

The procedure of backbone network follows the black lines, as illustrated in Figure \ref{fig:framework}.
For simplicity in presentation, we denote the visible-stream backbone network as function $F_{v}^{b}(\cdot)$ while $F_{t}^{b}(\cdot)$ for thermal-stream backbone network. Given a visible image $I_{v}$ and a thermal image $I_{t}$, the extracted features (\textbf{after the $BN$ layer}) from the backbone can be respectively calculated as,
\begin{align}
    v^b = F_{v}^{b}(I_{v}), \,\,\,\,\,\, t^b = F_{t}^{b}(I_{t}).
\end{align}
\subsection{Mid-level features incorporation}
\label{ssec:mid_feat}
As we all known, different CNN feature maps correspond to different semantic levels, from bottom to top layers, tending to visual concepts that are of higher semantic level and more abstract \cite{zeiler2014visualizing}. Therefore, we try to incorporate the mid-level features from middle layers of the CNN model through skip-connection to enhance the person features with more discriminability and robustness. We empirically adopt the features after $stage3$ as the mid-level features, then fuse them with those final features of backbone network.
As illustrated in Figure \ref{fig:framework}, the red lines depict the procedures for incorporating mid-level features.

The procedures of mid-level features incorporation for visible-stream and thermal-stream are all the same. For simplicity, here we take the visible-stream as an example to illustrate how to perform the mid-level features incorporation. Given the mid-level feature-maps $MLF3$ after $stage3$ of ResNet50 model and the final features $v^b$ (\textbf{before the $BN$ layer}) of backbone network from the visible-stream, similar to the backbone network, we sequentially fed the mid-level feature maps $MLF3$ into the global average pooling and full connected layers to obtain the mid-level features $v^m$.
The following is the most important step, how to fuse the mid-level features $v^m$ and final backbone features $v^b$ to obtain the final visible person features $v$.

For simplicity, we only consider the fusion mechanism to be summation ($sum$) or concatenation ($cat$).
\begin{align}
    sum: \,\,\,\,\,\,  v = v^m + v^b, \\
    cat: \,\,\,\,\,\,  v = [v^m; v^b].
\end{align}

Similarly, we also can obtain the final thermal person features $t$ by fusing the mid-level features $t^m$ and final backbone features $t^b$ from the thermal stream.
\subsection{Dual-modality triplet loss}
\label{ssec:d_trip}
After designing the network for feature extraction of visible and thermal images, we propose the dual-modality triplet loss to supervise the feature learning objectives, enhancing the discriminative person feature learning by focusing on both cross-modality discrepancy and intra-modality variations. It mainly consists of the cross-modality triplet loss and the intra-modality triplet loss.

\textbf{Triplet loss revisit.} Triplet loss is firstly proposed in FaceNet \cite{schroff2015facenet}. Given an anchor point $x_a$ with class label of $y_a$, triplet loss aims to make that the positive point $x_p$ belonging to the same class $y_a$ is closer to the anchor than that of a negative point belonging to another class $y_n$, by at least a margin $\rho$. Given some pre-selected triplets $\{x_a, x_ p, x_n\}$, the triplet loss can be represented as,
\begin{align}
    L_{tri} = \sum\limits_{\substack{a,p,n \\ y_a = y_p \neq y_n}} \left[\rho + \| x_a - x_p \|_2 - \| x_a - x_n \|_2\right]_+,
\end{align}
where $[x]_{+} = \max(x, 0)$ denotes the standard hinge loss.

For calculation simplicity and performance improving, Hermans et al. \cite{hermans2017defense} proposed an organizational modification to mine the hard triplets. The core idea is to form batches by randomly sampling $P$ identities, and then randomly sampling $K$ images of each identity, resulting in a mini-batch $PK$ images. For each sample $x_a$ in the mini-batch, we can select the hardest positive and hardest negative samples within the mini-batch to form the triplets for computing the batch hard triplet loss,
\begin{align}\label{eq:loss_bh}
    L_{bh\_tri}(X) = \overbrace{\sum\limits_{i=1}^{P} \sum\limits_{a=1}^{K}}^{\textnormal{all anchors}}
        \Big[\rho & + \hspace*{-5pt} \overbrace{\max\limits_{p=1 \dots K} \hspace*{-5pt} \|x^i_a - x^i_p\|_2 }^{\textnormal{hardest positive}} \\
               & - \hspace*{-5pt} \underbrace{\min\limits_{\substack{j=1 \dots P \\ n=1 \dots K \\ j \neq i}} \hspace*{-5pt} \| x^i_a - x^j_n \| _2}_{\textnormal{hardest negative}} \Big]_+,\nonumber
\end{align}
which is defined for a mini-batch $X$ and where a data point $x_i^j$ denotes the $j^{th}$ image of the $i^{th}$ person in the batch.

\textbf{Batch sampling method.} Due to our two-stream structure respectively extracting features for visible and thermal images, we introduce the following sampling strategy. Specially, $P$ person identities are firstly randomly selected at each iteration, and then we randomly select $K$ visible images and $K$ thermal images of the selected identity to form the mini-batch, in which totally $2*PK$ images. In our experiments, we set $P =  8$ and $K = 4$.

\textbf{Cross-modality triplet loss.} Based on the batch hard triplet loss, given the visible person features $V$ for $PK$ images and thermal person features $T$ for $PK$ images in a mini-batch, in a bi-directional manner the cross-modality triplet loss can be calculated as,
\begin{align}
\begin{tabular}{cl}
    $L_{c\_tri} (V,T) = \sum_{i=1}^{P} \sum_{a=1}^{K} \Big[ \rho $ & $+ \max\limits_{p=1,\cdots,K} \| v_a^{i} -  t_p^{i} \|_{2}$ \\
    & $- \min\limits_{\substack{j=1 \dots P \\ n=1 \dots K \\ j \neq i}} \| v_a^{i} -  t_n^{j} \|_{2} \Big]_{+}$
\end{tabular}, \nonumber
\end{align}
and
\begin{align}
\begin{tabular}{cl}
    $L_{c\_tri} (T,V) = \sum_{i=1}^{P} \sum_{a=1}^{K} \Big[ \rho $ & $+ \max\limits_{p=1,\cdots,K} \| t_a^{i} -  v_p^{i} \|_{2}$ \\
    & $- \min\limits_{\substack{j=1 \dots P \\ n=1 \dots K \\ j \neq i}} \| t_a^{i} -  v_n^{j} \|_{2} \Big]_{+}$
\end{tabular}. \nonumber
\end{align}
The overall cross-modality triplet loss is,
\begin{align}
    L_{c\_tri} = L_{c\_tri} (V,T) + L_{c\_tri} (T,V).   \label{eq:cross_trip}
\end{align}

\textbf{Intra-modality triplet loss.} Similarly, we also can compute the triplet loss on each modality.
\begin{align}
\begin{tabular}{cl}
    $L_{i\_tri} (V) = \sum_{i=1}^{P} \sum_{a=1}^{K} \Big[ \rho $ & $+ \max\limits_{p=1,\cdots,K} \| v_a^{i} -  v_p^{i} \|_{2}$ \\
    & $- \min\limits_{\substack{j=1 \dots P \\ n=1 \dots K \\ j \neq i}} \| v_a^{i} -  v_n^{j} \|_{2} \Big]_{+}$
\end{tabular},  \nonumber
\end{align}
and
\begin{align}
\begin{tabular}{cl}
    $L_{i\_tri} (T) = \sum_{i=1}^{P} \sum_{a=1}^{K} \Big[ \rho $ & $+ \max\limits_{p=1,\cdots,K} \| t_a^{i} -  t_p^{i} \|_{2}$ \\
    & $- \min\limits_{\substack{j=1 \dots P \\ n=1 \dots K \\ j \neq i}} \| t_a^{i} -  t_n^{j} \|_{2} \Big]_{+}$
\end{tabular}. \nonumber
\end{align}

The overall intra-modality triplet loss is,
\begin{align}
    L_{i\_tri} = L_{i\_tri} (V) + L_{i\_tri} (T). \label{eq:intra_trip}
\end{align}

\textbf{The dual-modality triplet loss.} Based on the cross-modality triplet loss (Eqn.(\ref{eq:cross_trip})) and intra-modality triplet loss (Eqn.(\ref{eq:intra_trip})), we can obtain the dual-modality triplet loss,
\begin{align}
    L_{d\_tri} = L_{c\_tri}  + \lambda_{1} L_{i\_tri} , \label{eq:dual_trip}
\end{align}
where $\lambda_{1}$ is a predefined trade-off parameters.

Moreover, similar to DCTR \cite{ye2018visible}, for the sake of feasibility and effectiveness for classification, the softmax loss is utilized to integrate the identity specific information by treating each person as a class. Therefore, the final loss is,
\begin{align}
    L_{all} = L_{softmax}  + \lambda_{2}  L_{d\_tri}, \label{eq:final_loss}
\end{align}
where $\lambda_{2}$ is a predefined trade-off parameters.

\section{Experiments}
\label{sec:exp}
In this section, we evaluate the effectiveness of our EDFL methods to enhance the person features for VT Re-ID tasks on two datasets, SYSU-MM01 \cite{wu2017rgb} and RegDB \cite{nguyen2017person}.
\subsection{Experimental settings}
\label{ssec:settings}
\textbf{Datasets and settings.} SYSU-MM01 \cite{wu2017rgb} is a large-scale dataset captured by 6 cameras, including 4 visible and 2 thermal cameras. Some cameras are deployed in the indoor environments and others are deployed in the outdoor environments. We adopt the single-shot all-search mode evaluation protocol predefined in \cite{wu2017rgb}, which is the most challenging setting according to the paper. The training set contains 395 persons, including 22258 visible images and 11909 thermal images. The testing set contains 96 persons, including 3803 thermal images for query and 301 randomly selected visible images as gallery set.

RegDB \cite{nguyen2017person} is constructed by dual camera systems, and includes 412 persons. For each person, 10 visible images are captured by a visible camera, and 10 thermal images are obtained by a thermal camera. We follow the evaluation protocol in \cite{ye2018hierarchical} and \cite{ye2018visible}, where the dataset is randomly split into two halves, one for training and the other for testing. For testing, the images from one modality (default is thermal) were used as the gallery set while the ones from the other modality (default is visible) as the probe set. The procedure is repeated for 10 trials to achieve statistically stable results, recording the mean values.

We adopt the cumulative matching characteristics (CMC) and the mean average precision (mAP) as the evaluation indicators to report the performances.
%The backbone flows (the black lines in Figure \ref{fig:framework}) only with the identity loss $L_{softmax}$ are utilized as the baseline of our methods.

\textbf{Features.} When without the mid-level features incorporation, the features after the final $BN$ layer in the backbone flows (the black lines in Figure \ref{fig:framework}) are adopted as the person representations during testing. While when incorporating the mid-level features, the features after the other final $BN$ layer in the skip-connection flows (the red lines in Figure \ref{fig:framework}) are adopted as the person representations during testing. Note that all the person features are $L2$ normalized for testing or for metric learning guided by our proposed dual-modality triplet loss.

\begin{table}
\small
\caption{Effectiveness of dual-modality triplet loss on the RegDB and SYSU-MM01 datasets. Re-identification rates (\%) at rank r and mAP (\%).}
\label{tab:d_tri}
  \centering
  \begin{tabular}{l|c|c|c|c}
    \toprule[2pt]
    RegDB  & r = 1 & r = 10 & r = 20 & mAP \\ \toprule[1pt]
     $L_{softmax}$ &\multirow{2}{*} {33.79} &  \multirow{2}{*} {55.05} & \multirow{2}{*} {65.68} &  \multirow{2}{*} {36.33}  \\
     (baseline) & & & & \\ \hline
     $L_{d\_tri}$ & 29.56 & 48.57 & 59.35 & 33.88   \\  \hline
     $L_{all}$ & \textbf{40.10} & \textbf{62.48} & \textbf{73.30} & \textbf{42.46}  \\ \toprule[1pt]\toprule[1pt]
     SYSU-MM01  & r = 1 & r = 10 & r = 20 & mAP \\ \toprule[1pt]
     $L_{softmax}$ & \multirow{2}{*} {26.81 } &  \multirow{2}{*} { 72.68} & \multirow{2}{*} {86.21 } & \multirow{2}{*} {31.13 }   \\
     (baseline) &  &  & &  \\ \hline
     $L_{d\_tri}$ & 31.20 & \textbf{78.30} & 88.49 &  \textbf{35.46}  \\  \hline
     $L_{all}$ & \textbf{31.45} & 77.61 & \textbf{88.74} & 35.39  \\\toprule[2pt]
  \end{tabular}
\end{table}

\textbf{Implementation details.} The implementation of our method is with PyTorch. We adopt the ResNet50 model pre-rained on ImageNet as the backbone network. The dimension of person features is set $d = 1024$. In training phase, the input images is resized to $288 \times 144$ and padded with 10, then randomly left-right flipped and cropped to $288 \times 144$ for data augmentation. We use the adam method \cite{kingma2014adam} as the optimizer with $\beta_1 = 0.9$ and $\beta_2 = 0.999$. The network is trained for 60 epochs for SYSU-MM01 and 30 epochs for RegDB. Those layers in the res-convolution blocks are fixed for 5 epochs firstly. The initial learning rate is set as 0.0001, and decayed with 0.1 at epoch 30. We set the predefined margin $\rho = 0.5$ for all the triplet losses.

\subsection{Ablation experiments}
\label{ssec:ablation}
We evaluate the effectiveness of our enhancing discriminative feature learning (EDFL) method, including two components, dual-modality triplet loss (DMTL) and mid-level features incorporation (MFI).
\subsubsection{The effectiveness of dual-modality triplet loss}
\label{sssec:exp_dtrip}
The proposed dual-modality triplet loss could be adopted with the identity softmax loss in the backbone network, the black lines in Figure \ref{fig:framework}. We adopt the identity softmax loss $L_{softmax}$ as our baseline method. Table \ref{tab:d_tri} lists the results with only identity softmax loss $L_{softmax}$, only our proposed dual-modality triplet loss $L_{d\_tri}$ and the full loss $L_{all}$ (Eqn.(\ref{eq:final_loss})), combination of $L_{softmax}$ and $L_{d\_tri}$.
From Table \ref{tab:d_tri}, we can find that.

\begin{figure}
\centering
\begin{tabular}{c@{\hspace{1mm}}c}
\includegraphics[width=40mm]{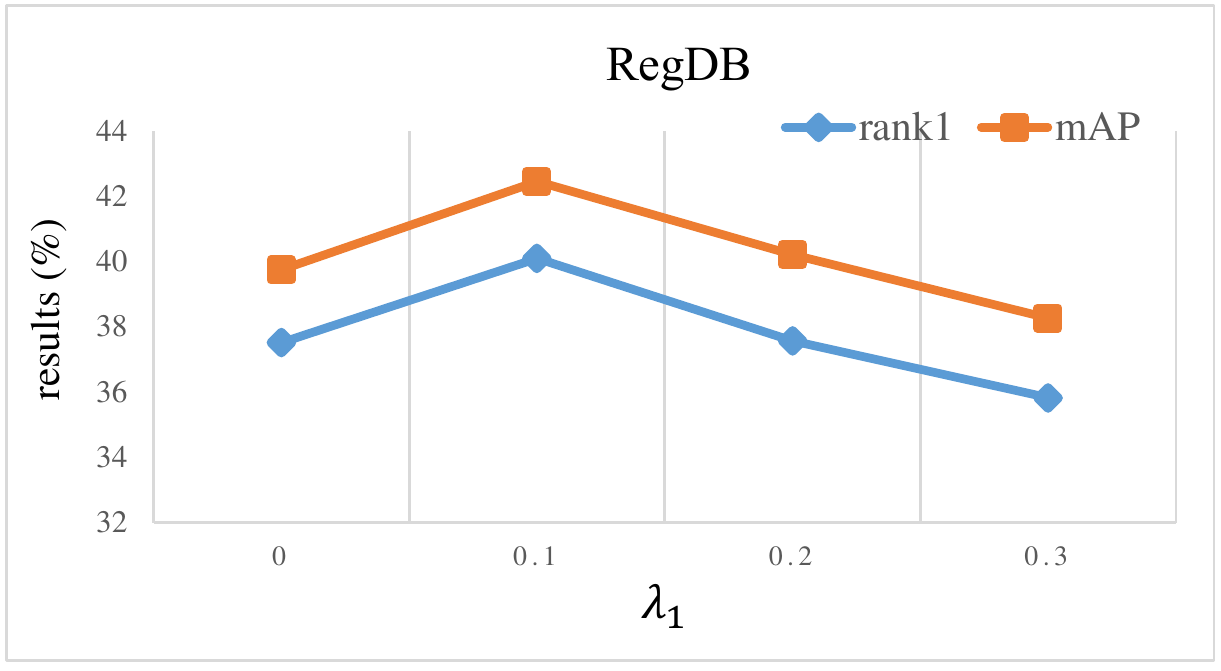} & \includegraphics[width=40mm]{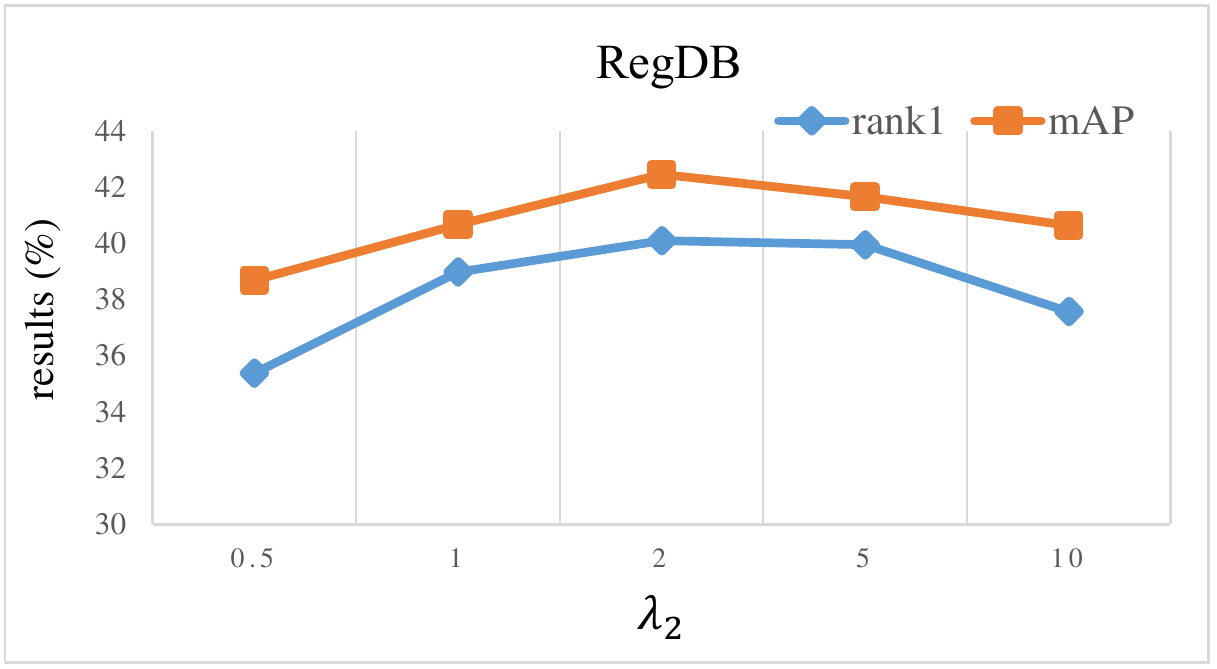} \\
(a) & (b) \\
\includegraphics[width=40mm]{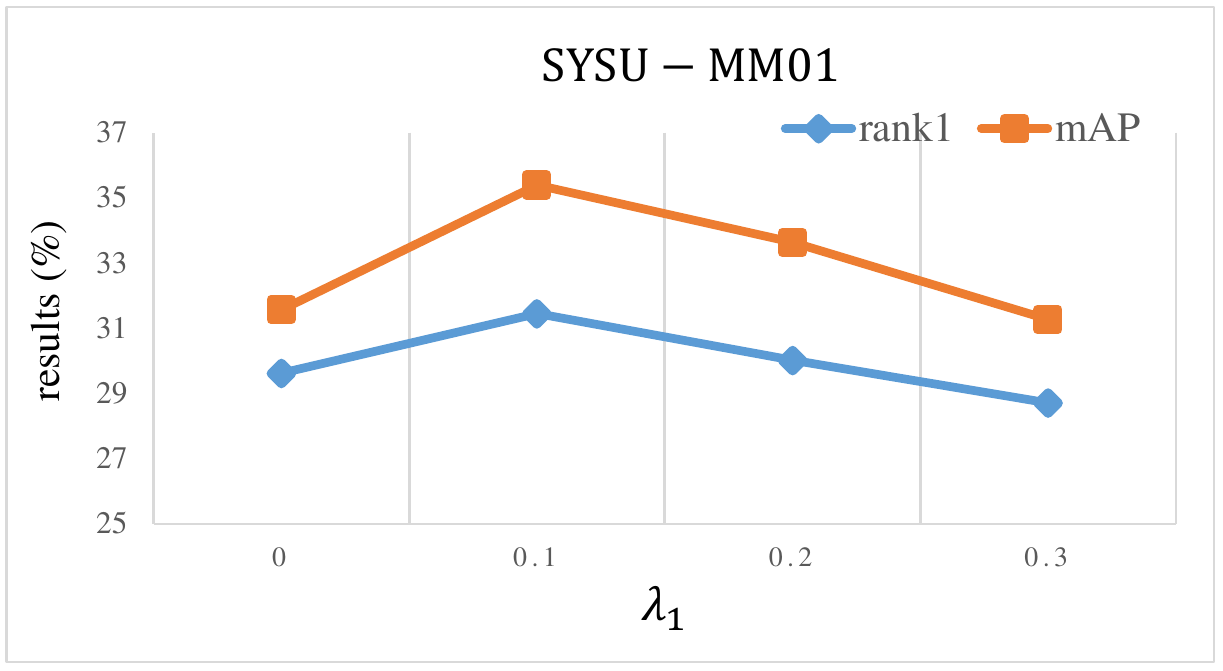} & \includegraphics[width=40mm]{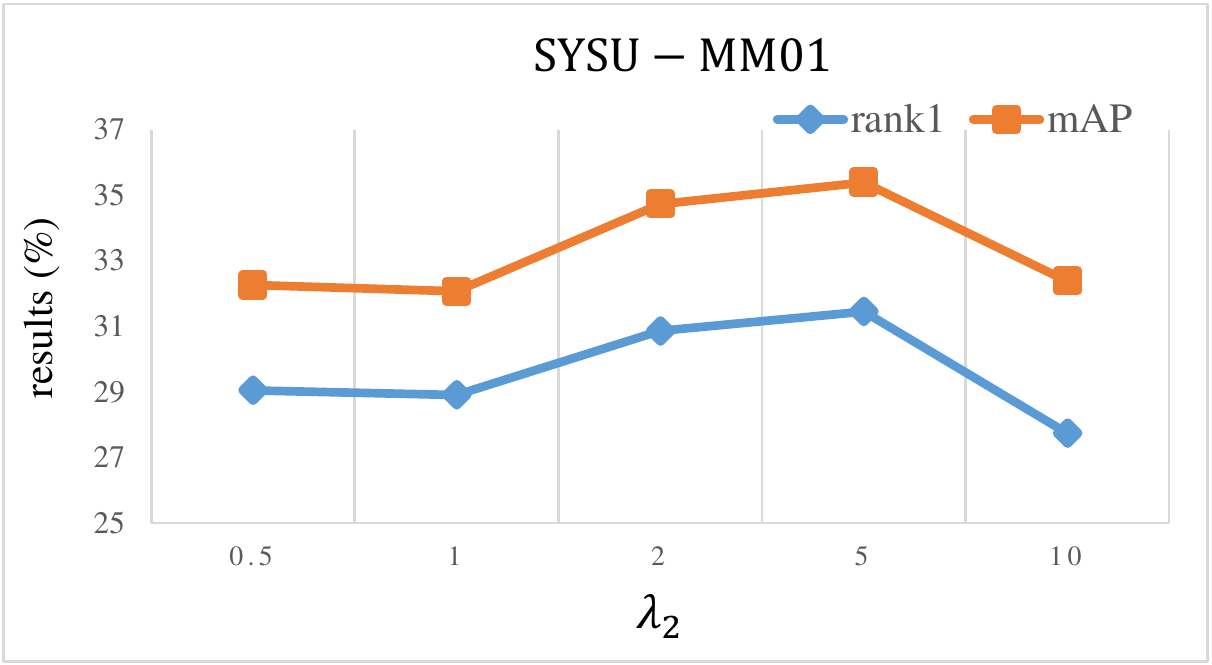} \\
(c) & (d)
\end{tabular}
\caption{The influence of two trade-off parameters $\lambda_1$ and $\lambda_2$ on the RegDB and SYSU-MM01 datasets. (a) The influence of $\lambda_1$ when set $\lambda_2 = 2$ on RegDB dataset. (b) The influence of $\lambda_2$ when set $\lambda_1 = 0.1$ on RegDB dataset. (c) The influence of $\lambda_1$ when set $\lambda_2 = 5$ on SYSU-MM01 dataset. (d) The influence of $\lambda_2$ when set $\lambda_1 = 0.1$ on SYSU-MM01 dataset.}
\label{fig:parameters}
\end{figure}

(1) On RegDB dataset, method with only our proposed dual-modality triplet loss $L_{d\_tri}$ performs worse compared to the baseline method with only identity softmax loss $L_{softmax}$. However, when combining the $L_{d\_tri}$ and $L_{softmax}$, the performances are drastically improved with a large margin.

(2) On SYSU-MM01 dataset, method with only our proposed dual-modality triplet loss $L_{d\_tri}$ performs better compared to the baseline method with only identity softmax loss $L_{softmax}$. The full loss $L_{all}$ achieves comparable performance to $L_{d\_tri}$.

(3) The different performance of $L_{d\_tri}$ on RegDB and SYSU-MM01 datasets, maybe come from the different size of the two datasets. RegDB is a small one wile SYSU-MM01 is bigger. The hard mining triplet loss is always performing well with large number of training samples.

(4) The improved performances on both of the RegDB and SYSU-MM01 datasets demonstrate the effectiveness of our proposed dual-modality triplet loss for guiding the training of neural networks to address the cross-modality discrepancy and intra-modality variations.

In our full loss $L_{all}$, there are two trade-off parameters, $\lambda_1$ for the intra-modality triplet loss in Eqn.(\ref{eq:dual_trip}) and $\lambda_2$ for the dual-modality triplet loss in Eqn.(\ref{eq:final_loss}). The influences of the two parameters are shown in Figure \ref{fig:parameters}.
For both of the RegDB and SYSU-MM01 datasets, when $\lambda_1 = 0.1$ it achieves the best performance, which denotes that with little intra-modality triplet loss truly could help to improve the cross-modality person Re-ID performance to some extent.
As to $\lambda_2$, it obtain the best performance when $\lambda_2 = 2$ on RegDB dataset, while $\lambda_2 = 5$ on SYSU-MM01 dataset.

\begin{table}
\small
\caption{Effectiveness of mid-level features incorporation on the RegDB and SYSU-MM01 datasets. Re-identification rates (\%) at rank r and mAP (\%).}
\label{tab:mfi}
  \centering
  \begin{tabular}{l|c|c|c|c}
    \toprule[2pt]
    RegDB  & r = 1 & r = 10 & r = 20 & mAP \\ \toprule[1pt]
     baseline & 33.79 &  55.05 & 65.68 &  36.33  \\ \hline
     $MFI3-512-sum$ & 27.52 & 42.67 & 53.50 & 29.60  \\
     $MFI3-512-cat$ & 43.96 & 63.67 & 73.23 & 46.94   \\ \toprule[1pt]
     $MFI3-1024-sum$ & 36.91 & 56.38 & 67.11 & 39.89  \\
     $MFI3-1024-cat$ & \textbf{49.82} & \textbf{66.93} & \textbf{75.97} & \textbf{51.06}   \\ \toprule[1pt]
     $MFI2-1024-cat$ & 42.89 & 60.81 & 73.82 & 44.54  \\  \hline
     $MFI3-1024-cat$ & \multirow{2}{*} {48.59 } & \multirow{2}{*} { 66.91} & \multirow{2}{*} { 75.12} &  \multirow{2}{*} {49.86 } \\
     (without $L_{backbone}$) &  &  &  &  \\ \toprule[1pt] \toprule[1pt]
     SYSU-MM01  & r = 1 & r = 10 & r = 20 & mAP \\ \toprule[1pt]
     baseline & 26.81  &  72.68  & 86.21  &  31.13   \\  \hline
     $MFI3-512-sum$ & 24.62 & 70.49 & 84.27 & 28.68   \\
     $MFI3-512-cat$ & 31.78 & 77.41 & 88.26 & 35.10   \\  \toprule[1pt]
     $MFI3-1024-sum$ & 29.98 & 74.98 & 86.52 & 33.05    \\
     $MFI3-1024-cat$ & 31.79 & 77.92 & \textbf{89.34} & 34.95    \\  \toprule[1pt]
     $MFI2-1024-cat$ & 29.93 & 73.46 & 86.82 &  32.97   \\ \hline
     $MFI3-1024-cat$ & \multirow{2}{*} { \textbf{32.91}} & \multirow{2}{*} {\textbf{77.95} } & \multirow{2}{*} { 88.97} &  \multirow{2}{*} { \textbf{35.17}}\\
     (without $L_{backbone}$) &  &  &  &  \\\toprule[2pt]
  \end{tabular}
\end{table}

\subsubsection{The effectiveness of mid-level features incorporation}
\label{sssec:exp_mlf}
To evaluate the effectiveness of mid-level features incorporation, based on the baseline method with only identity loss, we incorporate the mid-level features from $stage3$ (or $stage2$) with fusion methods $cat$ and $sum$, respectively. The dimension of person features is respectively set as 512 and 1024. The overall loss contains the loss of the backbone network $L_{backbone}$ (the black lines in Figure \ref{fig:framework}) and the loss of skip-connection branches (the red lines in Figure \ref{fig:framework}). The results are listed in Table \ref{tab:mfi}, where ``$MFI3-512-cat$'' denotes ``mid-level features from $stage3$ $-$ dimension of person feature representation $d = 512$ $-$ fusion method $cat$''.
From Table \ref{tab:mfi}, we can find that.

(1) For fusion methods, $cat$ performs much better than $sum$ in all the cases.

(2) With $cat$ fusion method, for the dimension of person feature representations, $d = 1024$ performs much better than $d = 512$ on RegDB dataset, while $d = 1024$ obtains comparable results to $d = 512$ on SYSU-MM01 dataset.

(3) For the mid-level features, $MFI2$ always performs worse than $MFI3$. It demonstrate that compared to $MFI2$, $MFI3$ can learn relative higher level person features which are important for recognition.

(4) For the loss of backbone network $L_{backbone}$, without it, a little worse results are obtained on RegDB dataset. However, without the $L_{backbone}$, a little better results are obtained on SYSU-MM01 datasets.

(5) Compared to the baseline, our proposed mid-level features incorporation methods perform much better with a large margin, demonstrating the effectiveness of our proposed MFI methods for enhancing the person features with more discrimination power.

\begin{table*}
\caption{Comparison to the state-of-the-arts on the RegDB and SYSU-MM01 datasets. Re-identification rates (\%) at rank r and mAP (\%).}
\label{tab:sota}
\centering
\begin{tabular}{l|c@{\hspace{6mm}}c@{\hspace{6mm}}c@{\hspace{3mm}}|@{\hspace{3mm}}c@{\hspace{3mm}}||c@{\hspace{6mm}}c@{\hspace{6mm}}c@{\hspace{3mm}}|@{\hspace{3mm}}c@{\hspace{3mm}}}
\toprule[2pt]
\multirow{2}{*} {Methods} & \multicolumn{4}{c||}{RegDB} & \multicolumn{4}{c}{SYSU-MM01} \\ \cline{2-9}
    & r = 1 & r = 10 & r = 20 & mAP & r = 1 & r = 10 & r = 20 & mAP \\ \hline
  Zero-Padding \cite{wu2017rgb} & 17.75 & 34.21 & 44.35 & 18.90 & 14.80 & 54.12 & 71.33 & 15.95  \\
  TONE \cite{ye2018hierarchical}& 16.87 & 34.03& 44.10 & 14.92 & 12.52 & 50.72 & 68.60 & 14.42  \\
  TONE + XQDA \cite{ye2018hierarchical}& 21.94 & 45.05 & 55.73 & 21.80 & 14.01 & 52.78 & 69.06 & 15.97  \\
  TONE + HCML \cite{ye2018hierarchical}& 24.44 & 47.53 & 56.78& 20.80 & 14.32 & 53.16 & 69.17 & 16.16  \\
  DCTR (BCTR) \cite{ye2018visible}& 32.67 & 57.64 & 66.58 & 30.99 & 16.12 & 54.90 & 71.47 & 19.15  \\
  DCTR (BDTR) \cite{ye2018visible}& 33.47 & 58.42 & 67.52 & 31.83 & 17.01 & 55.43 & 71.96 & 19.66  \\
  cmGAN \cite{dai2018cross} & - & - & - & - & 26.97 & 67.51 & 80.56 & 27.80  \\
  D$^2$RL \cite{wang2019learning1} &43.40 &66.10 &76.30 &44.10 &28.90 &70.60 &82.40 &29.20 \\ \hline \hline
  baseline & 33.79 & 55.05 & 65.68 & 36.33 & 26.81 & 72.68 & 86.21 & 31.13  \\
  DMTL (ours) & 40.10 & 62.48 & 73.30 & 42.46 & 31.45 & 77.61 & 88.74 & 35.39  \\
  MFI (ours) & 49.82 & 66.93 & 75.97 & 51.06 & 32.91 & 77.95 & 88.97 & 35.17  \\
  EDFL (ours) & \textbf{52.58} & \textbf{72.10} & \textbf{81.47} & \textbf{52.98} & \textbf{36.94} & \textbf{84.52} & \textbf{93.22} & \textbf{40.77}  \\ \hline \hline
  baseline + DCTR & 31.65 & 52.72 & 62.82 & 33.82 & 26.53 & 69.50 & 81.75 & 28.72  \\
  baseline + MFI + DCTR & 44.66 & 65.53 & 76.17 & 47.32 & 30.03 & 74.23 & 85.22 & 32.65  \\
\toprule[2pt]
\end{tabular}
\end{table*}

\subsubsection{The effectiveness of our enhancing discriminative feature learning method}
\label{sssec:exp_wdfl}

\begin{figure}
\centering
\includegraphics[width=80mm]{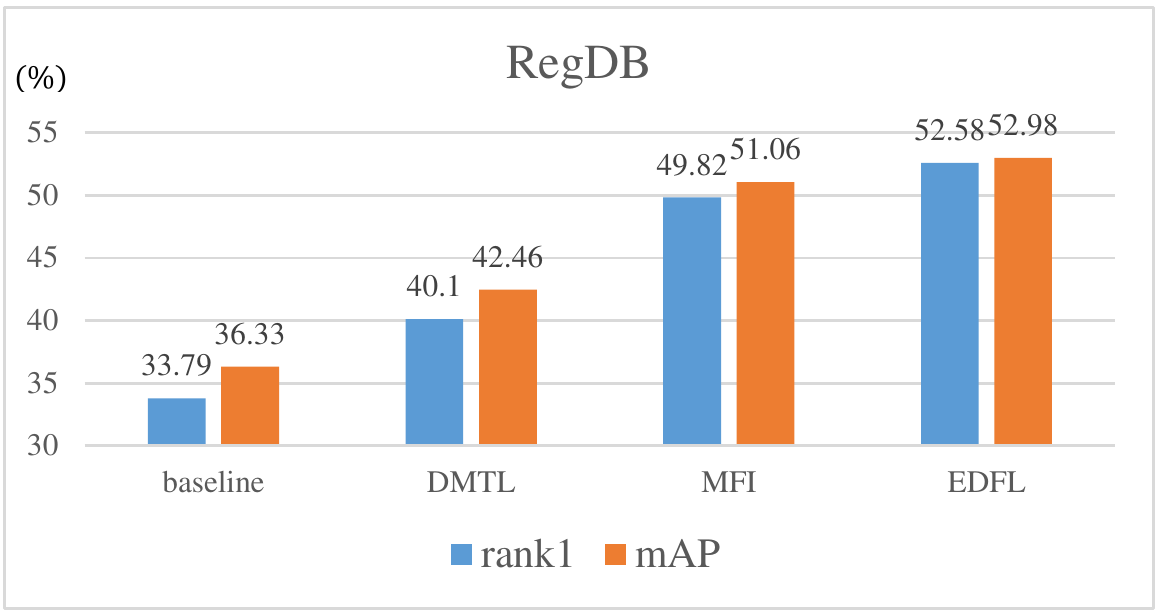} \\ (a)  \\
\includegraphics[width=80mm]{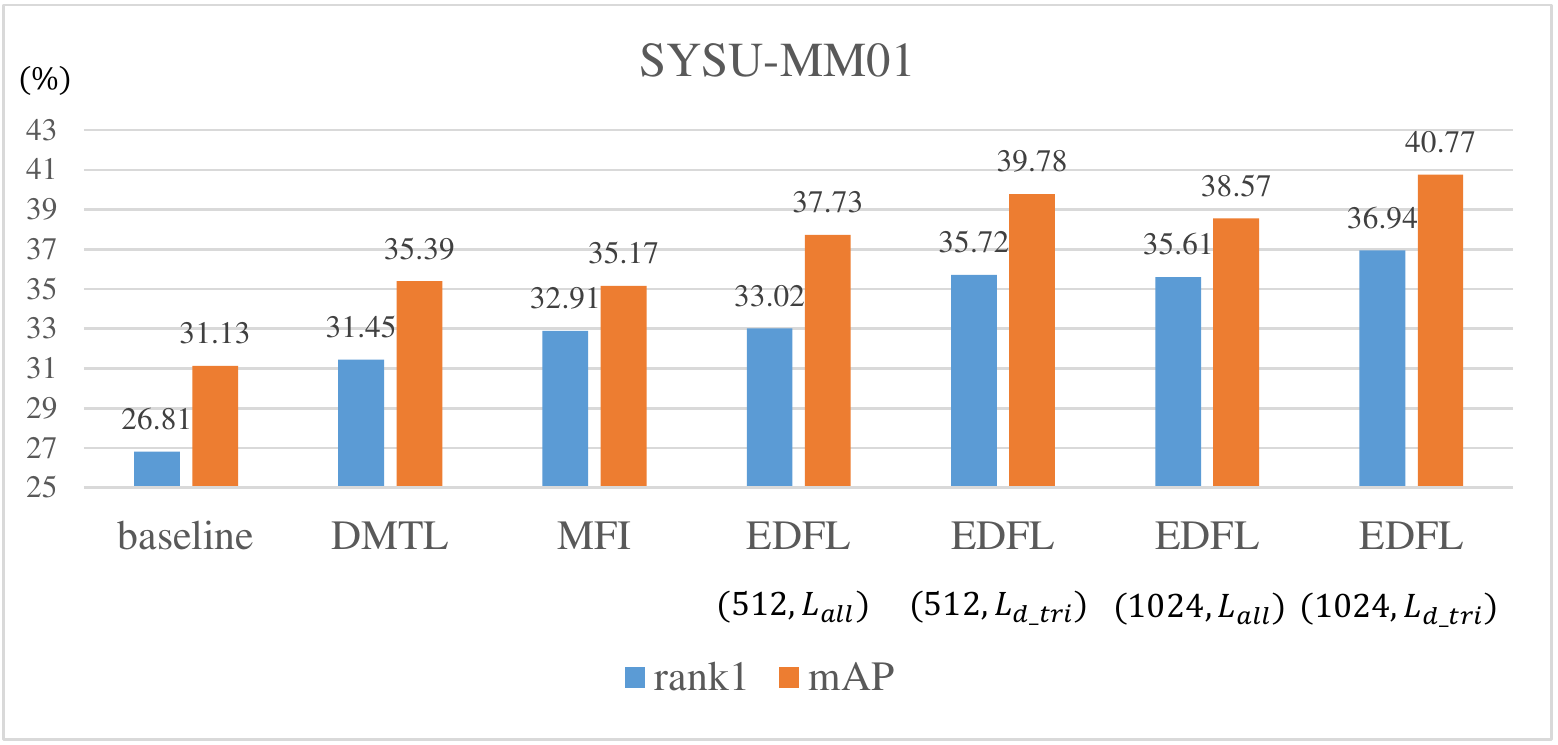} \\ (b)  \\
\caption{The performances of our proposed enhancing discriminative feature learning methods on (a) RegDB and (b) SYSU-MM01 datasets. For SYSU-MM01 dataset, we list the results of EDFL with different feature dimension ($d = 512 or 1024$), and different losses ($L_{all}$: both of the identity loss and our proposed dual-modality triplet loss, $L_{d\_tri}$: only our proposed dual-modality triplet loss).}
\label{fig:bdtr}
\end{figure}
Figure \ref{fig:bdtr} plots the final performance of our proposed EDFL methods including two components: DMTL and MFI.
On both of the RegDB and SYSU-MM01 datasets, our proposed DMTL and MFI methods respectively perform much better than the baseline method. Finally, our proposed EDFL method, combining the DMTL and MFI, achieves the best performance with a large margin.

\subsection{Comparison to the state-of-the-arts}
\label{ssec:exp_sota}

In this section, we compare our proposed EDFL with some state-of-the-art methods, zero-padding \cite{wu2017rgb}, HCML \cite{ye2018hierarchical}, DCTR \cite{ye2018visible}, cmGAN \cite{dai2018cross} and D$^2$RL \cite{wang2019learning1}.
Since \cite{wu2017rgb} and \cite{ye2018hierarchical} only evaluate their methods on one dataset, while \cite{ye2018visible} re-implemented them to evaluate on both datasets, Table \ref{tab:sota} lists the corresponding results originated from \cite{ye2018visible}.
We also adopt the DCTR loss into our EDFL framework to replace our DMTL loss for comparison.
From Table \ref{tab:sota} we can know that.

(1) Our proposed EDFL method outperforms the state-of-the-art method D$^2$RL \cite{wang2019learning1} by large margins (\%) on both of the RegDB and SYSU-MM01 datasets, Rank1: \textbf{9.18} (=52.58-43.40),  mAP:  \textbf{8.88} (=52.98-44.10) for RegDB dataset; Rank1: \textbf{8.04} (=36.94-28.90), mAP: \textbf{11.57} (=40.77-29.20) for SYSU-MM01 dataset. It demonstrates the effectiveness of our proposed EDFL method containing the dual-modality triplet loss and mid-level features incorporation to enhance the discriminative feature learning for VT Re-ID.

(2) Our baseline is better than DCTR (BDTR) \cite{ye2018visible}, even with a large margin on SYSU-MM01 dataset. The advantages maybe the following two folds: (i) the backbone networks and (ii) the sampling method.
(i) For the backbone network, we adopt ResNet50 \cite{he2016deep} while DCTR \cite{ye2018visible} adopts AlexNet \cite{krizhevsky2012imagenet}. ResNet model is more advanced in feature learning and image classification compared to AlexNet which is proposed almost 7 years ago.
(ii) For the sampling method, in each mini-batch, we randomly selected $P$ person identities and then randomly select $K$ visible images and $K$ thermal images of the selected identity, resulting mini-batch size $2*PK$. While DCTR \cite{ye2018visible} randomly selected $P$ person identities and then randomly select \textbf{only one} visible images and \textbf{only one} thermal images of the selected identity, resulting mini-batch size $2*P$. This sampling strategy leads that DCTR \cite{ye2018visible} can not mine the hardest positive samples compared to our sampling strategy. Actually, DCTR essentially does not do this. Moreover, we test our baseline with ResNet50 backbone and the sampling method adopted in DCTR \cite{ye2018visible}, obtaining the worse results (\%), Rank1: 25.49, mAP: 28.80 on RegDB dataset; Rank1: 23.95, mAP: 26.43 on SYSU-MM01 dataset.

(3) The baseline + DCTR performs worse than our DMTR, also the baseline + MFI + DCTR vs. our EDFL, where the main difference lies in the loss and sampling methods. The reason may be that DCTR loss is essentially bounded with the sampling strategy with $K=1$, which leads that DCTR can not mine the hardest positive samples, however, which is very important for addressing the cross-modality discrepancy and intra-modality variations. The results demonstrate the effectiveness of our DMTR loss in a hard triplets mining manner for VT Re-ID, compared to the DCTR loss in a contrastive top-ranking manner.

(4) Our EDFL, cmGAN \cite{dai2018cross} and  D$^2$RL \cite{wang2019learning}are all based on the ResNet50 for person feature extraction. cmGAN introduces the adversarial learning, and D$^2$RL introduces the extra image translation procedures.  However, our EDFL performs much better than cmGAN and D$^2$RL, without any auxiliary sub-tasks, demonstrating the effectiveness of the dual-modality triplet loss and mid-level features incorporation to enhance the discriminative feature learning for VT Re-ID.

\textbf{Different query settings.} We also evaluate the performance of different query settings on the RegDB dataset as done in \cite{ye2018visible}. The afore-reported results are under the default setting of ``Visible to Thermal'': visible images as query while thermal images as gallery. Here, we change the query to thermal images and gallery to visible images, ``Thermal to Visible''. EDFL achieves the results (\%), Rank1: 51.89, Rank10: 72.09, Rank20: 81.04, mAP: 52.13. The good performance may be attributed to the bi-directional calculation of our dual-modality triplet loss.

\section{Conclusion}
\label{sec:conclusion}
This work aims at enhancing the discriminative feature learning for VT Re-ID. We argue the simple ways to success. To address the cross-modality discrepancy and intra-modality variations, the proposed EDFL method mainly consists of two extreme simple components, mid-level features incorporation and the dual-modality triplet loss, focusing on improving the person features with more discriminability and guiding the training procedures. The experimental results with remarkable improvements demonstrate the effectiveness of the proposed EDFL method compared to the baseline and the state-of-the-arts.

\bibliographystyle{IEEEtrans}
\bibliography{reid}

\end{document}